\documentclass{article}

\usepackage{microtype}
\usepackage{graphicx}
\usepackage{subfigure}
\usepackage{booktabs} 

\usepackage{amsmath}
\usepackage{amssymb}

\usepackage{hyperref}

\usepackage[accepted]{icml2019}

\icmltitlerunning{A Personalized Affective Memory Neural Model for Improving Emotion Recognition}

\begin{document}

\twocolumn[
\icmltitle{A Personalized Affective Memory Model for Improving Emotion Recognition }




\begin{icmlauthorlist}
\icmlauthor{Pablo Barros}{uh}
\icmlauthor{German I. Parisi}{uh,ap}
\icmlauthor{Stefan Wermter}{uh}
\end{icmlauthorlist}

\icmlaffiliation{uh}{Knowledge Technology, Department of Informatics, University of Hamburg, Germany}
\icmlaffiliation{ap}{Apprente Inc, Montain View, CA, USA}

\icmlcorrespondingauthor{Pablo Barros}{barros@informatik.uni-hamburg.de}

\icmlkeywords{Emotion Recognition, Unsupervised Learning, Facial expressions}

\vskip 0.3in
]




\begin{abstract}
Recent models of emotion recognition strongly rely on supervised deep learning solutions for the distinction of general emotion expressions. However, they are not reliable when recognizing online and personalized facial expressions, e.g., for person-specific affective understanding. In this paper, we present a neural model based on a conditional adversarial autoencoder to learn how to represent and edit general emotion expressions. We then propose Grow-When-Required networks as personalized affective memories to learn individualized aspects of emotion expressions. Our model achieves state-of-the-art performance on emotion recognition when evaluated on \textit{in-the-wild} datasets. Furthermore, our experiments include ablation studies and neural visualizations in order to explain the behavior of our model.
\end{abstract}

\section{Introduction}
\label{introduction}
Automatic facial expression recognition became a popular topic in the past years due to the success of deep learning techniques. What was once the realm of specialists on adapting human-made descriptors of facial representations \cite{zhao2003face}, became one of many visual recognition tasks for deep learning enthusiasts \cite{schmidhuber2015deep}. The result is a continuous improvement in the performance of automatic emotion recognition during the last decade \cite{sariyanidi2015automatic} reflecting the progress of both hardware and different deep learning solutions.

However, the performance of deep learning models for facial expression recognition has recently stagnated \cite{soleymani2017survey, mehta2018facial}. The most common cause is the high dependence of such solutions on balanced, strongly labeled and diverse data. Recent approaches propose to address these problems by introducing transfer learning techniques \cite{ng2015deep, kaya2017video}, neural activation and data distribution regularization \cite{ding2017facenet2expnet, pons2018supervised}, and unsupervised learning of facial representations \cite{zen2014unsupervised, kahou2016emonets}. Most of these models present an improvement of performance when evaluated on specific datasets. Still, most of these solutions require strongly supervised training which may bias their capacity to generalize emotional representations. Recent neural models based on adversarial learning \cite{kim2017deep, saha2018unsupervised} overcome the necessity of strongly labeled data. Such models rely on their ability to learn facial representations in an unsupervised manner and present competitive performance on emotion recognition when compared to strongly supervised models.

Furthermore, it is very hard to adapt end-to-end deep learning-based models to different emotion recognition scenarios, in particular, real-world applications, mostly due to very costly training processes. As soon as these models need to learn a novel emotion representation, they must be re-trained or re-designed. Such problems occur due to understanding facial expression recognition as a typical computer vision task instead of adapting the solutions to specific characteristics of facial expressions \cite{adolphs2002recognizing}. 

We address the problem of learning adaptable emotional representations by focusing on improving emotion expression recognition based on two facial perception characteristics: the diversity of human emotional expressions and the personalization of affective understanding. A person can express happiness by smiling, laughing and/or by moving the eyes depending on who he or she is talking to, for example \cite{mayer1990perceiving}. The diversity of emotion expressions becomes even more complex when we take into consideration inter-personal characteristics such as cultural background, intrinsic mood, personality, and even genetics \cite{russell2017cross}. Besides diversity, the adaptation to learn personalized expressions is also very important. When humans already know a specific person they can adapt their own perception to how that specific person expresses emotions \cite{Hamann2004}. The interplay between recognizing an individual's facial characteristics, i.e. how they move certain muscles in their face, mouth and eye positions, blushing intensity, and clustering them into emotional states is the key for modeling a human-like emotion recognition system \cite{sprengelmeyer1998neural}.

The problem of learning diversity on facial expressions was addressed with the recent development of what is known as \textit{in-the-wild} emotion expression datasets \cite{zadeh2016multimodal, mollahosseini2017affectnet, dhall2018emotiw}. These corpora make use of a large amount of data to increase the variability of emotion representations. Although deep learning models trained with these corpora improved the performance on different emotion recognition tasks \cite{de2017face, kollias2018training}, they still suffer from the lack of adaptability to personalized expressions.

 Unsupervised clustering and dynamic adaptation were explored for solving the personalization of emotion expression problems \cite{chen2014linear, zen2014unsupervised, valenza2014revealing}. Most of the recent solutions use a collection of different emotion stimuli (EEG, visual and auditory, for example) and neural architectures, but the principle is the same: to train a recognition model with data from one individual person to improve affective representation. Although performing well in specific benchmark evaluations, these models are not able to adapt to online and continuous learning scenarios found on real-world applications. 
 
 The affective memory model \cite{barros2017self} addresses the problem of continuous adaptation by proposing a Growing-When-Required (GWR) network to learn emotional representation clusters from one specific person. One affective memory is created per person and it is populated with representations from the last layer of a convolutional neural network (CNN). This model creates prototypical neurons that represent each of the perceived facial expressions. Although improving emotion expression recognition when compared to state-of-the-art models, this model is not reliable on online scenarios. Prototype neurons will be formed only after a certain expression is perceived, and it relies heavily on a supervised CNN, so the model will only produce good performance once all the possible emotion expressions were perceived.
 
In this paper, we propose a personalized affective memory framework which improves the original affective memory model with respect to online emotion recognition. 
At the beginning of an interaction, humans tend to rely heavily on their own prior knowledge to understand facial expressions, with such prior knowledge learned over multiple episodes of interaction \cite{nook2015new}. We propose the use of a novel generative adversarial autoencoder model to learn the prior knowledge of emotional representations. We adapt the model to transfer the learned representations to unknown persons by generating edited faces with controllable emotion expressions \cite{huang2018high, wang2018facial}. We use the generated faces of a single person to initialize a personalized affective memory. We then update the affective memory with the perceived faces on an online manner and use it to recognize facial expressions online.
 
 We evaluate our model in two steps: first, how well the model can learn a general representation of emotion expressions, using the AffectNet \cite{mollahosseini2017affectnet} dataset. And second, how the novel personalized affective memory performs when recognizing emotion expressions of specific persons using the continuous expressions on the OMG-Emotion dataset \cite{BCLSSW18}. We compare our model with state-of-the-art solutions on both datasets. Besides performance alone, we also provide an exploratory  study on how our model works based on its neural activities using activation visualization techniques. 
 
\section{The Personalized Affective Memory}

\begin{figure*}[t]
\begin{center}
   \includegraphics[width=0.70\linewidth]{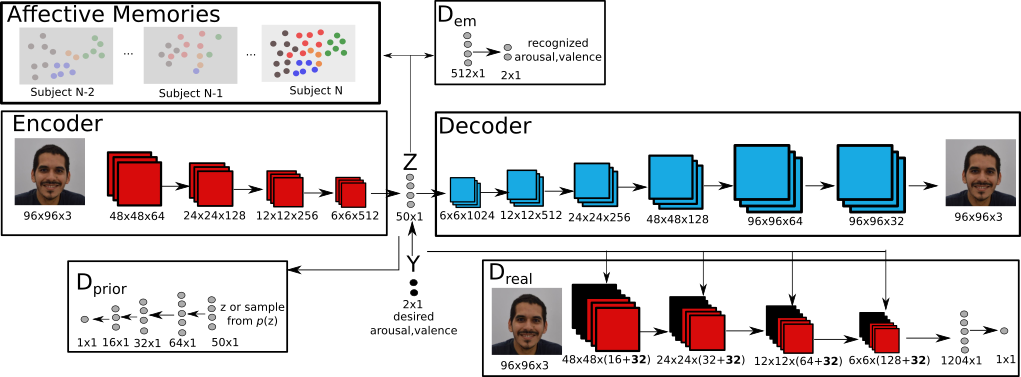}
\end{center}
   \caption{The P-AffMem model is composed of the Prior-Knowledge Adversarial autoencoder (PK) and the affective memories. The PK implements an encoder/decoder architecture and specific discriminators to detail arousal and valence ($D_{em}$), to ensure a prior distribution to the encoded representation ($D_{prior}$) and to guarantee a photo-realistic image generation ($D_{real}$). The affective memories are individual Growing-When-Required(GWR) networks which learn personalized aspects of the facial expression.}
\label{fig:paffmem}
\end{figure*}

The Personalized Affective Memory (P-AffMem) model is composed of two modules: the prior-knowledge learning module (PK) and the affective memory module. The prior-knowledge learning module consists of an adversarial autoencoder that learns facial expression representations ($z$) and generates images using controllable emotional information, represented as continuous arousal and valence values ($y$). 
After the autoencoder is trained, we use it to generate a series of edited images with different expressions from a person using different values of $y$. We use this collection of generated faces as initialization to a growing-when-required (GWR) network which represents the personalized affective memory for that specific person. We use the clusters of the personalized affective memory to recognize the emotion expressions of that person. Figure \ref{fig:paffmem} illustrates the topological details of the P-AffMem model. 

\subsection{Prior-Knowledge Adversarial Autoencoder}

Our PK model is an extended version of the Face Editing GAN \cite{lindt2019facial}, which was developed to learn how to represent and edit age information into facial images. We choose the Face Editing GAN as the basis for our prior-knowledge module due to its capability of working without paired-data (images of the same person with different ages) and to its robustness against statistical variations in the input data distribution.

The PK model consists of an encoder and generator architecture ($E$ and $G$), and three discriminator networks. The first one learns arousal/valence representations ($D_{em}$), the second guarantees that the facial expression representation has a uniform distribution ($D_{prior}$) and the third one assures that the generated image is photo-realistic and expresses the desired emotion ($D_{real}$). The model receives as input an image ($x$) and a continuous arousal/valence label ($y$). It produces a facial representation ($z$) and an edited image expressing the chosen arousal/valence ($x_{gen}$). 

\subsubsection{Encoder and Generator}

The encoder network ($E$) consists of four convolution layers and one fully connected output layer. We make use of a convolution stride of 2 to avoid the use of pooling, and thus learning its own spatial down-sampling \cite{radford2015unsupervised}. $E$ receives as input an RGB image with a dimension of $96x96x3$ and outputs the facial expression representation ($z$) with a dimension of ($50x1$). $z$ is concatenated with  the desired emotional representation ($y$) and  used as input to the decoder network ($G$) which generates an image ($x_{gen}$) with the desired emotional expression $y$.  The decoder is composed of six transposed convolution layers. The first four have a stride of 2, while the last two have a stride of 1. All the convolution and transposed convolution of $E$ and $G$ use a kernel size of 5x5. All the layers of $E$ and $G$ use ReLu activation functions. Both $E$ and $G$ are trained with an image reconstruction loss ($L{rec}$):

\begin{equation}
\label{caae_reconstruction_loss}
 \min_{E,G} L_{rec} = L_1 (x, G(E(x), z)) 
\end{equation}

where $L_1$ is the mean absolute error. 

One of the problems of the common GANS is the appearance of ghost artifacts on the reconstructed images. The Face Editing Gans authors address this problem by using a total variation minimization loss function \cite{mahendran2015understanding}, but with questionable results. We address this problem by using the identity-preserving loss ($L_{iden}$) used by the ExprGan \cite{ding2017exprgan} and the G2-GAN \cite{song2017geometry} on the reconstructed image. The identity preserving loss is computed using a pre-trained VGG face model \cite{parkhi2015deep}. The VGG face model proposes a topological update on the VGG16 model and is trained to identify persons based on their faces. To compute $L_{iden}$, we compare the activations of the five first convolutional layers of the VGG Face for $x$ and $x_{gen}$:

\begin{equation}
\label{iden_loss}
 \min_{E,G} L_{iden} = \sum\limits_{l} L_1(\phi_{l}(x), \phi_{l}(G(E(x), z))
\end{equation}

where $\phi_{l}$ is the activation of the $l$th layer of the VGG face. 

\subsubsection{ Arousal/Valence Representation Discriminator}

The arousal/valence representation discriminator ($D_{em}$) enforces the encoder to learn facial representations which are able to be used on an emotion recognition task. This discriminator is important to guarantee that $z$ is suitable to represent emotion expressions as $z$ will be used to populate the affective memory. $D_{em}$ consists of two fully connected hidden layers, the first one implementing a ReLu activation function and the last one a $tanh$ linear function. The last layer is used as output to represent continuous values of arousal and valence. The loss is calculated as the 

\begin{equation} 
\begin{aligned}
L_{em} = & \ \ \ \ \ MSE(D_{em}(x)^a, x^a) \\
 & + MSE(D_{em}(x)^v, x^v)
\end{aligned}
\end{equation}

 where MSE is the Mean-Squared error, $D_{em}(x)^a$ and $D_{em}(x)^v$ represent the arousal and valence output of $D_{em}$ respectively, and $x^a$ and $x^v$ the associated arousal and valence value to the image input $x$. 

\subsubsection{ Uniform Distribution Discriminator}

The uniform distribution discriminator ($D_{prior}$) is imposed on $z$ and enforces it to be uniformly distributed. It is important that $z$ is uniformly distributed to increase the generalization of the model when representing a facial expression. It receives as input $z$ itself or a randomly chosen sample from the uniform distribution $p_{prior}(z)$ and it has as goal to distinguish between them. $D_{prior}$ is composed of four fully connected layers, each one implementing a ReLu activation function. The adversarial loss function between $E$ and $D_{prior}$ is defined as: 
\begin{equation} 
\begin{aligned}
\min_E \max_{D_z} L_prior = & \ \ \ \ \ \mathbb{E}_{z_{prior} \sim p_{prior}(z)}[\log D_z(z_{prior})] \\
 & + \ \mathbb{E}_{x \sim p_{data}(x)}[\log (1 - D_z(E(x)))]
\end{aligned}
\end{equation}

where $E$ is the likelihood, \begin{math}p_{prior}(z)\end{math} the prior distribution imposed on the internal representation \begin{math}z\end{math} and \begin{math}p_{data}(x)\end{math} the distribution of the training images.

\subsubsection{Photo-realistic Image Discriminator}

The photo-realistic image discriminator ($D_{real}$) is basically a normal real/fake discriminator on the typical Generative Adversarial Network (GAN). In our case it also serves as a mechanism to impose the decoder ($G$) to produce photo-realistic images with the desired emotion expression ($y$). $D_{real}$ implements four convolution layers, with stride of 2 and kernel size of 5x5, followed by two fully connected layers. To enforce the desired arousal and valence ($y$) into the generated image ($x_{gen}$), we perform a re-shape and zero-padding operation to $y$ and concatenate it with each convolutional layer of $D_{real}$. The concatenation of $y$ to all the convolutional layers proved to be an important step in our experiments towards generating a photo-realistic edited facial expression. To enforce the desired $y$ into $x_{gen}$ produced by $G$, we use the following adversarial loss function: 

\begin{equation}\label{g_d_img_loss}
\begin{aligned}
\min_G \max_{D_{img}} L_{img} = & \ \ \ \ \mathbb{E}_{x,y \sim p_{data}(x,y)}[\log D_{img}(x,y)] \\
 & + \ \mathbb{E}_{x,y \sim p_{data}(x,y)} \\
& \ \ \ \ [\log (1 - D_{img}(G(E(x),y),y)]
\end{aligned}
\end{equation}

where $p_{data}(x,y)$ is the distribution of the training data.

\subsubsection{Overall Loss Function}

To train our model we use an overall loss-function defined as:

 \begin{equation} \label{total_loss}
 \begin{aligned}
 \min_{E, G} \max_{D_z, D_{img}} L_{total} = \lambda_1 L_{rec}+ \lambda_2 L_{em}+ \lambda_3 L_{iden} + \\  \lambda_4 L_{z}+ \lambda_5 L_{img}
 \end{aligned}
 \end{equation}

in which the coefficients \begin{math}\lambda_1\end{math}, \begin{math}\lambda_2\end{math}, \begin{math}\lambda_3\end{math}, \begin{math}\lambda_4\end{math} and \begin{math}\lambda_4\end{math} are used to balance the facial expression discrimination, the high-fidelity of the generated images and the presence of the desired emotion on the generated image.


\subsection{Affective Memory}


Growing-When-Required (GWR) networks \cite{Marsland2002} were deployed recently to address the problems of continuous learning \cite{Parisi2017, Parisi2018}. The capability of creating prototype neurons using an online manner allows the GWR to adapt quickly to changes on the input data. Which makes it ideal for our online learning mechanism.

Each neuron of the GWR consists of a weight vector ${w}_j$ representing prototypical information of the input data. A newly perceived emotion expression will be associated with a best-matching unit (BMU) $b$, which is calculated by minimizing the distances between the facial expression and all the neurons on the GWR. Given a set of $N$ neurons, $b$ with respect to the input ${x}\in\mathbb{R}^n$ is computed as:

\begin{equation}
\begin{aligned}
b = \arg\min_{j\in N} \left(\Vert {x} - {w}_j  \Vert^2 \right).
\end{aligned}
\end{equation}

New connections are created between the BMU and the second-BMU with relation to input.
When a BMU is computed, all the neurons the BMU is connected to are referred to as its topological neighbors. Each neuron is equipped with a habituation counter $h_i \in [0,1]$ expressing how frequently it has been fired based on a simplified model of how the efficacy of a habituating synapse reduces over time.

The habituation rule is given by $\Delta h_i=\tau_i \cdot \kappa \cdot (1-h_i)-\tau_i$, where $\kappa$ and $\tau_i$ are constants that control the decreasing behavior of the habituation counter \cite{Marsland2002}. To establish whether a neuron is habituated, its habituation counter $h_i$ must be smaller than a given habituation threshold $t_h$.

The network is initialized with two neurons and, at each learning iteration, it inserts a new neuron whenever the activity of the network ($a(i) = \exp(-\left(\Vert {x} - {w}_j  \Vert^2 \right))$, when an expression $i$ is perceived, of a habituated neuron is smaller than a given threshold $t_a$, i.e., a new neuron is created if $a(i)<t_a$ and $h_i<t_h$. 

The P-AffMem model uses the GWR as affective memories to perform personalization learning of prototype neurons. The GWR is created when the first facial expression of a person is perceived. The PK module generates 200 edited images with combinations of arousal and valence within the interval of [-1, 1], with increments of 0.01. These images are used to initialize the affective memory and act as a transfer knowledge from the PK initial estimations. To avoid that the generated samples dominate the affective memory update over time, we stop generating samples after a certain number of expressions were perceived.

To guarantee that the PK does not dominate the training of the affective memory over time, we use a novel update function to modulate the impact of the PK on the GWR. The training of the network is carried out by adapting the BMU according to:

\begin{equation}\label{eq:UpdateRateW}
\Delta \textbf{w}_j = \epsilon_i \cdot h_i \cdot (\textbf{x} - \textbf{w}_j) \cdot (1-a(i)),
\end{equation}

where $\epsilon_i$ is a constant learning rate. If real faces are perceived, and they are different from the ones encoded by the PK, the activation of the network will be smaller and the impact on the weight update will be higher. The affective memory will be encouraged to create new neurons to represent newly perceived expressions instead of the ones coming from the PK.

To allow the GWR to perform classification of emotion expressions, we implement an associative labeling \cite{Parisi2017} to each neuron. During the training phase, we assign to each neuron two continuous values, representing arousal and valence. When training with an example that comes from the generated images, we update the labels of the BMU using the desired arousal and valence. The update is modulated by a labeling learning factor $\gamma$ which is defined during training. To categorize a newly perceived expression, we just read the labels of the BMU associated with it. 

\subsection{Parameter Optimization}

We trained the PK model using $L_{total}$ and optimized the model's topology, coefficients, the batch size and the training parameters using a TPE optimization strategy \cite{bergstra2013hyperopt}. The optimization was based on two characteristics: the objective performance on the arousal/valence discriminator and the minimization of $L_{total}$. The final coefficient values are as follows:
\begin{math}\lambda_1=1\end{math},
\begin{math}\lambda_2=0.02\end{math}, \begin{math}\lambda_3=0.3\end{math}, \begin{math}\lambda_4=0.01\end{math} and \begin{math}\lambda_5=0.01\end{math}. The batch size is 48. The normal distribution with mean 0 and standard deviation 0.02 is employed for the initialization of the weights of all layers. All biases are initially set to 0.
For optimization, the Adam Optimizer with learning rate \begin{math}\alpha=0.0002\end{math}, \begin{math}\beta_1=0.5\end{math} and \begin{math}\beta_2=0.999\end{math} is employed.

\begin{table}[h]
\caption{Training parameters of each affective memory.}
    \center \begin{tabular}{ c|  c }
        Parameter & Value \\
	\hline
    Epochs & 10 \\
    Activity threshold ($t_a$) & 0.4 \\    
    Habituation threshold ($t_h$) & 0.2\\
    Habituation modulation ($\tau_i$ and $\kappa$) & 0.087, 0.032\\
    Labeling factor ($\gamma$) & 0.4
    \end{tabular} 
\label{tab:affMemParam}
\end{table}

We also use TPE optimization to tune the parameters of the affective memory. Although they were created individually for one specific person, all of them have the same hyperparameters to simplify our evaluation task and allow the model to be used online. Given that the GWR adapts to the input data distribution, and we maintain the same data nature, we do not believe that fine-tuning each GWR to each individual person would obtain a large gain on recognition performance. Table \ref{tab:affMemParam} displays the final affective memory parameters.

\section{Experimental Setup}

To evaluate and understand better the individual impact of the PK and the affective memory on the recognition of emotion expressions, we perform two different types of experiments. First, we run a series of ablation studies to asses the contribution of each mechanism on the PK for general emotion recognition. Second, we run an emotion recognition experiment with the entire framework to assess the impact of the personalization on the emotion recognition performance.

\subsection{Datasets}

The AffectNet dataset \cite{mollahosseini2017affectnet} is composed of more than 1 million images with facial expressions collected from different \textit{in-the-wild} sources, such as Youtube and Google Images. More than 400 thousands images were manually annotated with continuous arousal and valence. The dataset has its own separation between training, testing and validation samples. The labels for the testing samples are not available, thus, all our experiments are performed using the training and validation samples. The large labelled data distribution of the AffectNet is important to guarantee that the PK learns general emotion recognition. It provides an ideal corpus to asses the impact of each proposed mechanism in the PK using an objectively comparable measure.


Furthermore, the AffectNet does not discriminate between images from the same person, so personalization does not play an important role. To evaluate the contributions of personalization, we provide final experiments on the One-Minute Gradual-Emotional Behavior dataset (OMG-Emotion) \cite{BCLSSW18}. The OMG-Emotion is composed of Youtube videos which are about one minute long and are annotated taking into consideration a continuous emotional behavior. The videos were selected using a crawler technique that uses specific keywords based on long-term emotional scenes such as "monologues", "auditions", "dialogues" and "emotional scenes", which guarantees that each video has only one person performing an emotional display. A total of 675 videos were collected, which sums up to around 10h of data. Each utterance on the videos is annotated with two continuous labels, representing arousal and valence. The emotion expressions displayed in the OMG-Emotion dateset are heavily impacted by person-specific characteristics which are highlighted by the gradual change of emotional behavior over the entire video.

\subsection{Experiments}

Our experiments are divided into two categories: the ablation studies (A) and the personalized emotion recognition (P). The A category is divided into a series of experiments, each of them evaluating one of the discriminators on the PK: the photo-realistic image discriminator $(D_{real})$, the uniform distribution ($D_{prior}$) and the arousal/valence discriminator($D_{em}$). We train the PK in each of these experiments with the training subset of the AffectNet corpus and evaluate it using the validation subset.

We first train the PK without any of the previously mentioned discriminators ($PK_{base}$) to guarantee an unbiased baseline. Then, we repeat the training, now adding each discriminator individually. We  report experimental results by combining the presence of each discriminator. Finally, we add all the discriminators ($PK_{all}$) and train the PK again. 

To provide a standard evaluation metric, we use the encoder representation ($z$) as input to an emotion recognition classifier. The classifier is composed of the same topology as the arousal/valence discriminator, and it is post-trained using the same training subset of the AffectNet dataset. When training the PK with the arousal/valence discriminator, we do not use the emotion recognition classifier, so that we can assess the performance of this specific discriminator. 

The P category is divided into two experiments: first, the PK is pre-trained with the AffectNet dataset and used to evaluate the test set of the OMG-Emotion dataset. Then we use the entire P-AffMem framework with the affective memories and repeat the experiment, now using one affective memory for each video of the test set. During our optimization routine we found that if the PK generates faces for more than 1s of the videos, the affective memory did improve the final results. As the OMG-Emotion dataset has a framerate of 25 frames per second, we turn off the PK image generation after the first 25 frames.

To evaluate the dimensional arousal and valence recognition, we use the Concordance Correlation Coefficient (CCC)~\cite{lawrence1989concordance} between the outputs of the model and the true labels. The CCC can be computed as:

\begin{equation}
CCC = \frac{2 \rho \sigma_x \sigma_y}{\sigma_{x}^2 + \sigma_{y}^2 + (\mu_x - \mu_y)^2}
\label{eq:ccc}
\end{equation}
where $\rho$ is the Pearson's Correlation Coefficient between model prediction labels and the annotations, $\mu_x$ and $\mu_y$ denote the mean for model predictions and the annotations and $\sigma_{x}^2$ and $\sigma_{y}^2$ are the corresponding variances.

\section{Results}
\subsection{Ablation Studies (A)}

Our ablation studies, summarized in Table \ref{tab:AExperiments}, help us to understand and quantify the impact of each of the PK modules on recognizing emotion expressions. The baseline model, without any extra discriminator, can be evaluated as a simple Generative Adversarial Network (GAN). Without surprise, the baseline model did not perform well on emotion recognition, achieving the worst result in our experiments. This reflects the inability of the baseline model to provide a good facial expression discrimination capability. Evaluating the model by introducing each discriminator allows us to asses their impact. The emotion recognition discriminator ($D_{em}$) seems to be the one which impacts most on the PK performance, which not surprising, as it enforces the PK encoder to learn specific characteristics for facial expression recognition. Nevertheless, we also observe a strong impact on the photo-realistic discriminator ($D_{real}$), indicating that the network benefits greatly from enforcing the encoding and generation of characteristics that discriminate between individuals. The prior distribution discriminator ($D_{prior}$) has a smaller individual contribution than all the others. However, when combined with the other discriminators, it provides a great improvement on the recognition. 

\begin{table}[h]
\caption{Concordance Correlation Coefficient (CCC) for arousal and valence when evaluating the different discriminator of the PK on the validation subset of the AffectNet dataset.}
    \center \begin{tabular}{ | c|  c | c |}
    \hline    
    Model & Arousal & Valence \\
    \hline
    $PK_{base}$ & 0.03 & 0.05 \\
    $PK_{base}$ + $D_{prior}$ & 0.03 & 0.08 \\
    $PK_{base}$ + $D_{real}$ & 0.08 & 0.09 \\
    $PK_{base}$ + $D_{em}$ & 0.18 & 0.22 \\
    $PK_{base}$ + $D_{prior}$ + $D_{real}$ & 0.11 & 0.18 \\
    $PK_{base}$ + $D_{prior}$ + $D_{em}$ & 0.25 & 0.35 \\
    $PK_{base}$ + $D_{real}$ + $D_{em}$ & 0.29 & 0.45 \\
    $PK_{all}$   & 0.38 & 0.67 \\ \hline
    \cite{mollahosseini2017affectnet}   & 0.34 & 0.60 \\ \hline
    \end{tabular} 
\label{tab:AExperiments}
\end{table}

Training the PK with all the discriminators yield the best results. To the best of our knowledge, the only reported CCC for arousal and valence on the AffectNet corpus comes from the authors of the dataset themselves \cite{mollahosseini2017affectnet}. They use an AlexNet convolutional neural network \cite{krizhevsky2012imagenet}  re-trained to recognize arousal and valence. Our PK model presents a better general performance improving the CCC in more than 0.04 for arousal and 0.07 for arousal. As our ablation studies are only intended to shed light on the impact of our PK modules, we did not pursue a pure benchmark study with other datasets.

\subsection{Personalized Recognition (P)}

we perform the experiments with and without the affective memory in order to quantify its contribution to the P-AffMem model. Table \ref{tab:PExperiments} summarizes the achieved CCC and the current state-of-the-art results on the OMG-Emotion dataset. The presence of the affective memories greatly improve the performance of the model with an increase on the achieved CCC of 0.9 and 0.7 for arousal and valence, respectively, when compared to the PK with all discriminators. 

The P-AffMem achieved the best results up to this point on the OMG-Emotion dataset. This dataset was recently used as part of a challenge and different solutions, mostly based on pre-trained deep learning models, were presented. Our model achieved a CCC improvement of 0.07 on arousal and 0.04 in valence, when compared to the winner of the challenge \cite{zheng2018multimodal}, which made use of audio/visual processing. The same occurred with the second best model \cite{peng2018deep}. The best model which used only facial expressions \cite{deng2018multimodal} achieved an arousal and valence CCC which were 0.16 and 0.18 smaller than our model. Further results are available at the OMG-Emotion website \footnote{https://bit.ly/2XUJBjq}.

\begin{table}[h]
\caption{Concordance Correlation Coefficient (CCC) for arousal and valence when evaluating our model on the testing subset of the OMG-Emotion dataset.}
    \center \begin{tabular}{ | c|  c | c |}
    \hline    
    Model & Arousal & Valence \\
    \hline
    $PK_{base}$ & -0.06 & -0.10 \\
    $PK_{base}$ + $D_{prior}$ & 0.02 & 0.01 \\
    $PK_{base}$ + $D_{real}$ & 0.04 & 0.02 \\
    $PK_{base}$ + $D_{em}$ & 0.09 & 0.12 \\
    $PK_{base}$ + $D_{prior}$ + $D_{real}$ & 0.13 & 0.13 \\
    $PK_{base}$ + $D_{prior}$ + $D_{em}$ & 0.21 & 0.29 \\
    $PK_{base}$ + $D_{real}$ + $D_{em}$ & 0.27 & 0.36 \\
    $PK_{all}$   & 0.32 & 0.46 \\ 
    $P-AffMem$   & 0.43 & 0.53 \\ \hline
    \cite{zheng2018multimodal}   & 0.36 & 0.49 \\
    \cite{peng2018deep}  & 0.24 & 0.43 \\
    \cite{deng2018multimodal}   & 0.27 & 0.35 \\ \hline
    \end{tabular} 
\label{tab:PExperiments}
\end{table}

\section{Discussions}


The PK module must be pre-trained to allow a better emotion expression recognition generalization. With 16 million parameters, the PK model has an extensive and computational expensive training procedure, as common to any adversarial network training. Our hypothesis was that, once trained, the PK would provide a general emotion recognition capability. Our experiments with the OMG-Emotion dataset demonstrate that the PK achieved a performance similar to other pre-trained deep learning models on recognizing general emotion expressions.

The affective memory contributes to the P-AffMem by providing a self-adaptable mechanism that improves the performance of the PK by clustering the individual characteristics of how a person expresses emotions. This assumption was demonstrated objectively as the P-AffMem has a higher CCC than the PK alone.

To extend our understanding of the P-AffMem and to pinpoint our claims on general versus specific emotion recognition, we will discuss in the next sessions the individual contributions of both the PK and the affective memories. We also present an insight into how the models contribute to the performance increase. 

\subsection{The Impact of the Discriminator Mechanisms}

\begin{figure}
\begin{center}
   \includegraphics[width=0.9\linewidth]{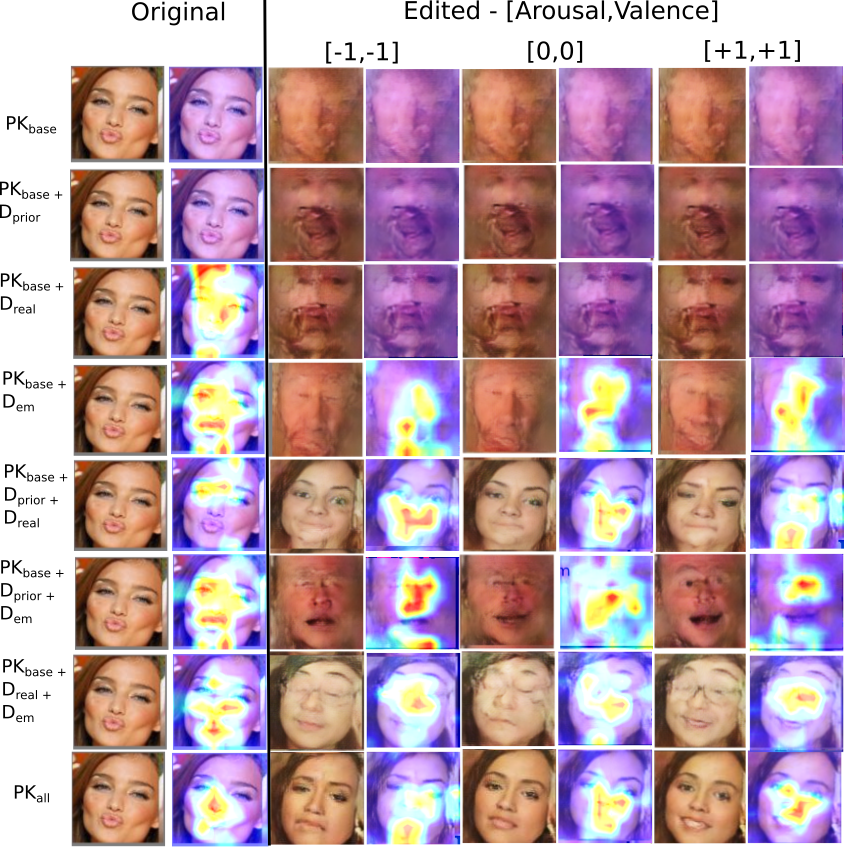}
\end{center}
   \caption{The first pair of the column represents the perceived emotion expression and the neural activation map of the last layer of the encoder. The remaining pairs of columns display the edited face using the indicated arousal and valence value, and its respective neural activation map.}
\label{fig:affNetDistribution}
\end{figure}

As an adversarial autoencoder, the PK is trained to both encode and edit a facial image. Each discriminator imposes a specific characteristic to both the encoding and editing tasks, and together they showed to be optimal, in our setup, for emotion expression recognition. While the impact of the discriminators on encoding a facial expression was objectively measured by our experiments, its contributions to face editing were just implicitly evaluated. By demonstrating that the affective memory improved the performance of the model, we guarantee that the edited faces used to initialize and train the affective memory actually have the desired facial expressions.

As the model is optimized to recognize facial expressions, an analysis of the pixel-wise quality of the edited faces can be misleading. To illustrate this assumption, we fed one image to each of the PK combinations proposed in our A category experiments and produced three edited faces with the extreme values of arousal and valence [-1,-1] and [1,1], and a neutral value [0,0]. We then calculated the mean neural activation maps \cite{zhou2016learning} for the last convolution layers of the discriminator for each of these examples. Figure \ref{fig:affNetDistribution} exhibits four pairs of columns, where the first is composed of the original image fed to the PKs and the obtained encoder activation visualizations. The other tree columns exhibit the edited images with the indicated arousal and valence values and the encoder activation. As the training of the encoder is directly affected by the discriminators, the visualizations give us an indication on how each discriminator impacts the learned representations, and help us to explain the network objective performance.

Clearly, the encoder trained only with $PK_{base}$ does not learn any meaningful information, which is reflected in the edited images and activation. This is also backed up by our objective results. The combination of the discriminators impact the edited faces and encoder in a distinguishable manner. While $PK_{real}$ enforces a realistic characteristic on the edited faces, it does impact the encoder on focusing facial structures which are not reliable for face expression representation. This effect translates on a lower CCC when compared to $D_{em}$, for example. When $D_{em}$ is present, the encoder clearly filtered facial expression-rich regions, such as eyes and mouth regions. This is clearly perceptible in the $PK_{base} + PK_{prior} + PK_{photo}$ combination, which presents a distinguishable facial editing but lower CCC than $PK_{base} + D_{real} + D_{em}$, for example. The encoder activation of the former indicates that it ignores some facial characteristics, which could explain the lower performance.

Finally, the combination of all discriminators gave the PK the best objective performance, and this reflects on an edited image and activation. The images contain present clear facial features, without the formation of distorting artifacts, maintaining the general facial proportions and characteristics. The activation shows that the encoder clearly focuses on the eyes-to-mouth region most of the times. These visualizations demonstrate the reliability of the PK in both encoding and editing facial expressions, which is very important for the optimal functioning of the affective memory.

\subsection{Affective Memory Behavior}

\begin{figure}[h]
\begin{center}
   \includegraphics[width=0.9\linewidth]{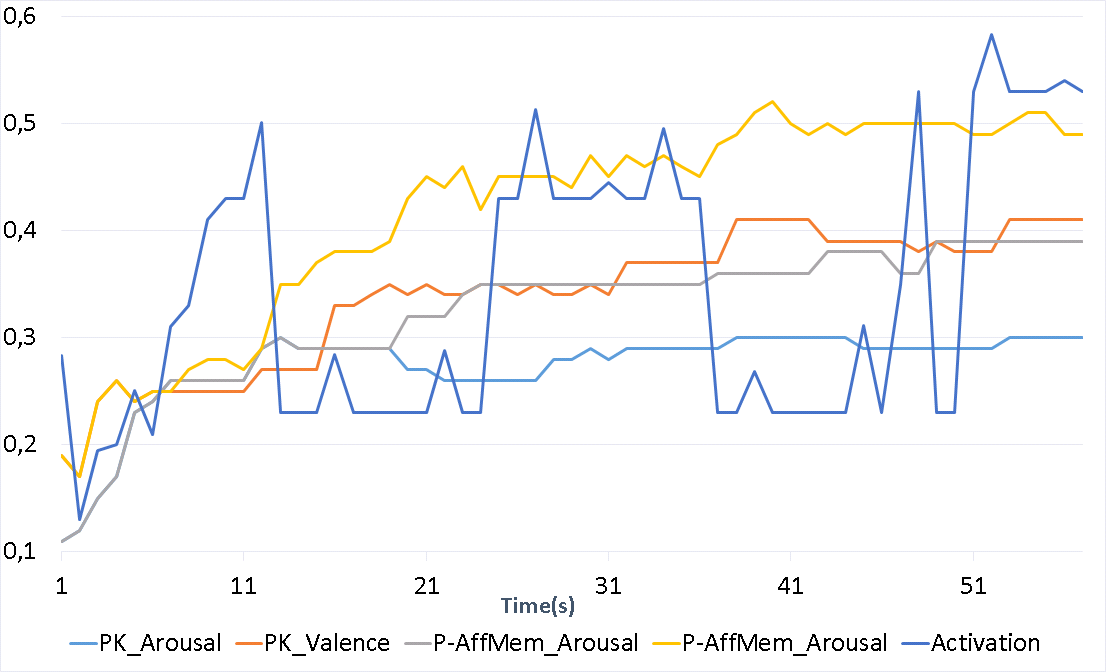}
\end{center}
   \caption{Performance of the model, measured in CCC for both PK and P-AffMem and neural activation when processing the "fd41c38b2" video of the OMG-Emotion dataset.}
\label{fig:activationDisplay}
\end{figure}

The videos on the OMG-Emotion dataset are heavily impacted by personalization characteristics, as they contain one person performing a monologue-like act for more than 1 minute. As the P-AffMem is initialized with the images edited by the PK, it contains, from the beginning, prototype neurons with a wide range of associated arousal and valence labels. As the video goes on, the neurons of the affective memory are much more influenced by newly perceived facial expressions which make the neural activation increase. Important to note that a newly perceived expression does not change the neuron's labels, only its prototypical information. This way, we guarantee that the network maintains a reliable classification.

As the PK already gives the affective memory general prototype neurons with reliable labels, the neural update makes them much more accurate towards specific characteristics of the newly perceived expressions. After some seconds have passed, the affective memory starts to perform better than the PK. This effect is demonstrated in Figure \ref{fig:activationDisplay}, which illustrates the performance of the PK and the P-AffMem, measured as arousal and valence CCC, and the neural activity evolution of the affective memory when processing the video "fd41c38b2" of the OMG-Emotion dataset. 

\section{Conclusion and Future Work}

The development of our Personalized Affective Memory (P-AffMem) model was inspired by two facial perception characteristics: the understanding of generalized emotional concepts and the online adaptation of individualized aspects of facial expressions. The general prior knowledge (PK) adversarial autoencoder was trained to transfer facial characteristics, represented as arousal and valence, to unknown persons. The affective memories contributed to the learned representations by creating prototypical representations of emotions for a specific person in an online fashion. Our evaluation demonstrated that our model achieved state-of-the-art performance on facial expression recognition, and we presented different insights on how, and why, our model works.




A clear limitation of our model is the processing of instantaneous emotion expressions. To address this problem, recent work on recurrent self-organization and sequence generation would be encouraged~\cite{Parisi2017, Parisi2018}. Another direction which could be explored is the integration of multisensory information, but with particular care for asynchronous affective perception, e.g., from prosody in speech and language understanding.

\section*{Acknowledgments}
The authors gratefully acknowledge partial support from the German Research Foundation DFG under project CML (TRR 169).






\bibliography{egbib}

\begin{thebibliography}{46}
\providecommand{\natexlab}[1]{#1}
\providecommand{\url}[1]{\texttt{#1}}
\expandafter\ifx\csname urlstyle\endcsname\relax
  \providecommand{\doi}[1]{doi: #1}\else
  \providecommand{\doi}{doi: \begingroup \urlstyle{rm}\Url}\fi

\bibitem[Adolphs(2002)]{adolphs2002recognizing}
Adolphs, R.
\newblock Recognizing emotion from facial expressions: psychological and
  neurological mechanisms.
\newblock \emph{Behavioral and cognitive neuroscience reviews}, 1\penalty0
  (1):\penalty0 21--62, 2002.

\bibitem[Barros \& Wermter(2017)Barros and Wermter]{barros2017self}
Barros, P. and Wermter, S.
\newblock A self-organizing model for affective memory.
\newblock In \emph{Neural Networks (IJCNN), 2017 International Joint Conference
  on}, pp.\  31--38. IEEE, 2017.

\bibitem[Barros et~al.(2018)Barros, Churamani, Lakomkin, Siqueira, Sutherland,
  and Wermter]{BCLSSW18}
Barros, P., Churamani, N., Lakomkin, E., Siqueira, H., Sutherland, A., and
  Wermter, S.
\newblock The omg-emotion behavior dataset.
\newblock In \emph{Proceedings of the International Joint Conference on Neural
  Networks (IJCNN 2018)}, pp.\  1408--1414. IEEE, Jul 2018.
\newblock \doi{arXiv:1803.05434v2}.

\bibitem[Bergstra et~al.(2013)Bergstra, Yamins, and Cox]{bergstra2013hyperopt}
Bergstra, J., Yamins, D., and Cox, D.~D.
\newblock Hyperopt: A python library for optimizing the hyperparameters of
  machine learning algorithms.
\newblock In \emph{Proceedings of the 12th Python in Science Conference}, pp.\
  13--20. Citeseer, 2013.

\bibitem[Chen et~al.(2014)Chen, Wang, Yang, and Chen]{chen2014linear}
Chen, Y.-A., Wang, J.-C., Yang, Y.-H., and Chen, H.
\newblock Linear regression-based adaptation of music emotion recognition
  models for personalization.
\newblock In \emph{Acoustics, Speech and Signal Processing (ICASSP), 2014 IEEE
  International Conference on}, pp.\  2149--2153. IEEE, 2014.

\bibitem[de~Bittencourt~Zavan et~al.(2017)de~Bittencourt~Zavan, Gasparin,
  Batista, e~Silva, Albiero, Bellon, and Silva]{de2017face}
de~Bittencourt~Zavan, F.~H., Gasparin, N., Batista, J.~C., e~Silva, L.~P.,
  Albiero, V., Bellon, O. R.~P., and Silva, L.
\newblock Face analysis in the wild.
\newblock In \emph{Graphics, Patterns and Images Tutorials (SIBGRAPI-T), 2017
  30th SIBGRAPI Conference on}, pp.\  9--16. IEEE, 2017.

\bibitem[Deng et~al.(2018)Deng, Zhou, Pi, and Shi]{deng2018multimodal}
Deng, D., Zhou, Y., Pi, J., and Shi, B.~E.
\newblock Multimodal utterance-level affect analysis using visual, audio and
  text features.
\newblock \emph{arXiv preprint arXiv:1805.00625}, 2018.

\bibitem[Dhall et~al.(2018)Dhall, Kaur, Goecke, and Gedeon]{dhall2018emotiw}
Dhall, A., Kaur, A., Goecke, R., and Gedeon, T.
\newblock Emotiw 2018: Audio-video, student engagement and group-level affect
  prediction.
\newblock In \emph{Proceedings of the 2018 on International Conference on
  Multimodal Interaction}, pp.\  653--656. ACM, 2018.

\bibitem[Ding et~al.(2017{\natexlab{a}})Ding, Sricharan, and
  Chellappa]{ding2017exprgan}
Ding, H., Sricharan, K., and Chellappa, R.
\newblock Exprgan: Facial expression editing with controllable expression
  intensity.
\newblock \emph{arXiv preprint arXiv:1709.03842}, 2017{\natexlab{a}}.

\bibitem[Ding et~al.(2017{\natexlab{b}})Ding, Zhou, and
  Chellappa]{ding2017facenet2expnet}
Ding, H., Zhou, S.~K., and Chellappa, R.
\newblock Facenet2expnet: Regularizing a deep face recognition net for
  expression recognition.
\newblock In \emph{Automatic Face \& Gesture Recognition (FG 2017), 2017 12th
  IEEE International Conference on}, pp.\  118--126. IEEE, 2017{\natexlab{b}}.

\bibitem[Hamann \& Canli(2004)Hamann and Canli]{Hamann2004}
Hamann, S. and Canli, T.
\newblock Individual differences in emotion processing.
\newblock \emph{Current opinion in neurobiology}, 14\penalty0 (2):\penalty0
  233--238, 2004.

\bibitem[Huang et~al.(2018)Huang, Chen, Wu, Lin, and Suganthan]{huang2018high}
Huang, B., Chen, W., Wu, X., Lin, C.-L., and Suganthan, P.~N.
\newblock High-quality face image generated with conditional boundary
  equilibrium generative adversarial networks.
\newblock \emph{Pattern Recognition Letters}, 111:\penalty0 72--79, 2018.

\bibitem[Kahou et~al.(2016)Kahou, Bouthillier, Lamblin, Gulcehre, Michalski,
  Konda, Jean, Froumenty, Dauphin, Boulanger-Lewandowski,
  et~al.]{kahou2016emonets}
Kahou, S.~E., Bouthillier, X., Lamblin, P., Gulcehre, C., Michalski, V., Konda,
  K., Jean, S., Froumenty, P., Dauphin, Y., Boulanger-Lewandowski, N., et~al.
\newblock Emonets: Multimodal deep learning approaches for emotion recognition
  in video.
\newblock \emph{Journal on Multimodal User Interfaces}, 10\penalty0
  (2):\penalty0 99--111, 2016.

\bibitem[Kaya et~al.(2017)Kaya, G{\"u}rp{\i}nar, and Salah]{kaya2017video}
Kaya, H., G{\"u}rp{\i}nar, F., and Salah, A.~A.
\newblock Video-based emotion recognition in the wild using deep transfer
  learning and score fusion.
\newblock \emph{Image and Vision Computing}, 65:\penalty0 66--75, 2017.

\bibitem[Kim et~al.(2017)Kim, Yoo, Kwak, Choi, and Kim]{kim2017deep}
Kim, Y., Yoo, B., Kwak, Y., Choi, C., and Kim, J.
\newblock Deep generative-contrastive networks for facial expression
  recognition.
\newblock \emph{arXiv preprint arXiv:1703.07140}, 2017.

\bibitem[Kollias \& Zafeiriou(2018)Kollias and Zafeiriou]{kollias2018training}
Kollias, D. and Zafeiriou, S.
\newblock Training deep neural networks with different datasets in-the-wild:
  The emotion recognition paradigm.
\newblock In \emph{2018 International Joint Conference on Neural Networks
  (IJCNN)}, pp.\  1--8. IEEE, 2018.

\bibitem[Krizhevsky et~al.(2012)Krizhevsky, Sutskever, and
  Hinton]{krizhevsky2012imagenet}
Krizhevsky, A., Sutskever, I., and Hinton, G.~E.
\newblock Imagenet classification with deep convolutional neural networks.
\newblock In \emph{Advances in neural information processing systems}, pp.\
  1097--1105, 2012.

\bibitem[Lawrence \& Lin(1989)Lawrence and Lin]{lawrence1989concordance}
Lawrence, I. and Lin, K.
\newblock A concordance correlation coefficient to evaluate reproducibility.
\newblock \emph{Biometrics}, pp.\  255--268, 1989.

\bibitem[Lindt et~al.(2019)Lindt, Barros, Siqueira, and
  Wermter]{lindt2019facial}
Lindt, A., Barros, P., Siqueira, H., and Wermter, S.
\newblock Facial expression editing with continuous emotion labels.
\newblock In \emph{2019 14th IEEE International Conference on Automatic Face \&
  Gesture Recognition (FG 2019)}, pp.\  1--8. IEEE, 2019.

\bibitem[Mahendran \& Vedaldi(2015)Mahendran and
  Vedaldi]{mahendran2015understanding}
Mahendran, A. and Vedaldi, A.
\newblock Understanding deep image representations by inverting them.
\newblock In \emph{Proceedings of the IEEE conference on computer vision and
  pattern recognition}, pp.\  5188--5196, 2015.

\bibitem[Marsland et~al.(2002)Marsland, Shapiro, and Nehmzow]{Marsland2002}
Marsland, S., Shapiro, J., and Nehmzow, U.
\newblock A self-organising network that grows when required.
\newblock \emph{Neural Networks}, 15\penalty0 (8--9):\penalty0 1041--1058,
  2002.

\bibitem[Mayer et~al.(1990)Mayer, DiPaolo, and Salovey]{mayer1990perceiving}
Mayer, J.~D., DiPaolo, M., and Salovey, P.
\newblock Perceiving affective content in ambiguous visual stimuli: A component
  of emotional intelligence.
\newblock \emph{Journal of personality assessment}, 54\penalty0 (3-4):\penalty0
  772--781, 1990.

\bibitem[Mehta et~al.(2018)Mehta, Siddiqui, and Javaid]{mehta2018facial}
Mehta, D., Siddiqui, M. F.~H., and Javaid, A.~Y.
\newblock Facial emotion recognition: A survey and real-world user experiences
  in mixed reality.
\newblock \emph{Sensors}, 18\penalty0 (2):\penalty0 416, 2018.

\bibitem[Mollahosseini et~al.(2017)Mollahosseini, Hasani, and
  Mahoor]{mollahosseini2017affectnet}
Mollahosseini, A., Hasani, B., and Mahoor, M.~H.
\newblock Affectnet: A database for facial expression, valence, and arousal
  computing in the wild.
\newblock \emph{arXiv preprint arXiv:1708.03985}, 2017.

\bibitem[Ng et~al.(2015)Ng, Nguyen, Vonikakis, and Winkler]{ng2015deep}
Ng, H.-W., Nguyen, V.~D., Vonikakis, V., and Winkler, S.
\newblock Deep learning for emotion recognition on small datasets using
  transfer learning.
\newblock In \emph{Proceedings of the 2015 ACM on international conference on
  multimodal interaction}, pp.\  443--449. ACM, 2015.

\bibitem[Nook et~al.(2015)Nook, Lindquist, and Zaki]{nook2015new}
Nook, E.~C., Lindquist, K.~A., and Zaki, J.
\newblock A new look at emotion perception: Concepts speed and shape facial
  emotion recognition.
\newblock \emph{Emotion}, 15\penalty0 (5):\penalty0 569, 2015.

\bibitem[Parisi et~al.(2018)Parisi, Tani, Weber, and Wermter]{Parisi2018}
Parisi, G., Tani, J., Weber, C., and Wermter, S.
\newblock Lifelong learning of spatiotemporal representations with dual-memory
  recurrent self-organization.
\newblock In \emph{arXiv:1805.10966}, 2018.

\bibitem[Parisi et~al.(2017)Parisi, Tani, Weber, and Wermter]{Parisi2017}
Parisi, G.~I., Tani, J., Weber, C., and Wermter, S.
\newblock Lifelong learning of humans actions with deep neural network
  self-organization.
\newblock \emph{Neural Networks}, 96:\penalty0 137--149, 2017.

\bibitem[Parkhi et~al.(2015)Parkhi, Vedaldi, Zisserman, et~al.]{parkhi2015deep}
Parkhi, O.~M., Vedaldi, A., Zisserman, A., et~al.
\newblock Deep face recognition.
\newblock In \emph{BMVC}, number~3, pp.\ ~6, 2015.

\bibitem[Peng et~al.(2018)Peng, Zhang, Ban, Fang, and Winkler]{peng2018deep}
Peng, S., Zhang, L., Ban, Y., Fang, M., and Winkler, S.
\newblock A deep network for arousal-valence emotion prediction with
  acoustic-visual cues.
\newblock \emph{arXiv preprint arXiv:1805.00638}, 2018.

\bibitem[Pons \& Masip(2018)Pons and Masip]{pons2018supervised}
Pons, G. and Masip, D.
\newblock Supervised committee of convolutional neural networks in automated
  facial expression analysis.
\newblock \emph{IEEE Transactions on Affective Computing}, 9\penalty0
  (3):\penalty0 343--350, 2018.

\bibitem[Radford et~al.(2015)Radford, Metz, and
  Chintala]{radford2015unsupervised}
Radford, A., Metz, L., and Chintala, S.
\newblock Unsupervised representation learning with deep convolutional
  generative adversarial networks.
\newblock \emph{arXiv preprint arXiv:1511.06434}, 2015.

\bibitem[Russell(2017)]{russell2017cross}
Russell, J.~A.
\newblock Cross-cultural similarities and differences in affective processing
  and expression.
\newblock In \emph{Emotions and Affect in Human Factors and Human-Computer
  Interaction}, pp.\  123--141. Elsevier, 2017.

\bibitem[Saha et~al.(2018)Saha, Navarathna, Helminger, and
  Weber]{saha2018unsupervised}
Saha, S., Navarathna, R., Helminger, L., and Weber, R.~M.
\newblock Unsupervised deep representations for learning audience facial
  behaviors.
\newblock \emph{arXiv preprint arXiv:1805.04136}, 2018.

\bibitem[Sariyanidi et~al.(2015)Sariyanidi, Gunes, and
  Cavallaro]{sariyanidi2015automatic}
Sariyanidi, E., Gunes, H., and Cavallaro, A.
\newblock Automatic analysis of facial affect: A survey of registration,
  representation, and recognition.
\newblock \emph{IEEE transactions on pattern analysis and machine
  intelligence}, 37\penalty0 (6):\penalty0 1113--1133, 2015.

\bibitem[Schmidhuber(2015)]{schmidhuber2015deep}
Schmidhuber, J.
\newblock Deep learning in neural networks: An overview.
\newblock \emph{Neural networks}, 61:\penalty0 85--117, 2015.

\bibitem[Soleymani et~al.(2017)Soleymani, Garcia, Jou, Schuller, Chang, and
  Pantic]{soleymani2017survey}
Soleymani, M., Garcia, D., Jou, B., Schuller, B., Chang, S.-F., and Pantic, M.
\newblock A survey of multimodal sentiment analysis.
\newblock \emph{Image and Vision Computing}, 65:\penalty0 3--14, 2017.

\bibitem[Song et~al.(2017)Song, Lu, He, Sun, and Tan]{song2017geometry}
Song, L., Lu, Z., He, R., Sun, Z., and Tan, T.
\newblock Geometry guided adversarial facial expression synthesis.
\newblock \emph{arXiv preprint arXiv:1712.03474}, 2017.

\bibitem[Sprengelmeyer et~al.(1998)Sprengelmeyer, Rausch, Eysel, and
  Przuntek]{sprengelmeyer1998neural}
Sprengelmeyer, R., Rausch, M., Eysel, U.~T., and Przuntek, H.
\newblock Neural structures associated with recognition of facial expressions
  of basic emotions.
\newblock \emph{Proceedings of the Royal Society of London B: Biological
  Sciences}, 265\penalty0 (1409):\penalty0 1927--1931, 1998.

\bibitem[Valenza et~al.(2014)Valenza, Citi, Lanat{\'a}, Scilingo, and
  Barbieri]{valenza2014revealing}
Valenza, G., Citi, L., Lanat{\'a}, A., Scilingo, E.~P., and Barbieri, R.
\newblock Revealing real-time emotional responses: a personalized assessment
  based on heartbeat dynamics.
\newblock \emph{Scientific reports}, 4:\penalty0 4998, 2014.

\bibitem[Wang et~al.(2018)Wang, Li, Mu, Huang, and Wang]{wang2018facial}
Wang, X., Li, W., Mu, G., Huang, D., and Wang, Y.
\newblock Facial expression synthesis by u-net conditional generative
  adversarial networks.
\newblock In \emph{Proceedings of the 2018 ACM on International Conference on
  Multimedia Retrieval}, pp.\  283--290. ACM, 2018.

\bibitem[Zadeh et~al.(2016)Zadeh, Zellers, Pincus, and
  Morency]{zadeh2016multimodal}
Zadeh, A., Zellers, R., Pincus, E., and Morency, L.-P.
\newblock Multimodal sentiment intensity analysis in videos: Facial gestures
  and verbal messages.
\newblock \emph{IEEE Intelligent Systems}, 31\penalty0 (6):\penalty0 82--88,
  2016.

\bibitem[Zen et~al.(2014)Zen, Sangineto, Ricci, and Sebe]{zen2014unsupervised}
Zen, G., Sangineto, E., Ricci, E., and Sebe, N.
\newblock Unsupervised domain adaptation for personalized facial emotion
  recognition.
\newblock In \emph{Proceedings of the 16th international conference on
  multimodal interaction}, pp.\  128--135. ACM, 2014.

\bibitem[Zhao et~al.(2003)Zhao, Chellappa, Phillips, and
  Rosenfeld]{zhao2003face}
Zhao, W., Chellappa, R., Phillips, P.~J., and Rosenfeld, A.
\newblock Face recognition: A literature survey.
\newblock \emph{ACM computing surveys (CSUR)}, 35\penalty0 (4):\penalty0
  399--458, 2003.

\bibitem[Zheng et~al.(2018)Zheng, Cao, Chen, and Xu]{zheng2018multimodal}
Zheng, Z., Cao, C., Chen, X., and Xu, G.
\newblock Multimodal emotion recognition for one-minute-gradual emotion
  challenge.
\newblock \emph{arXiv preprint arXiv:1805.01060}, 2018.

\bibitem[Zhou et~al.(2016)Zhou, Khosla, Lapedriza, Oliva, and
  Torralba]{zhou2016learning}
Zhou, B., Khosla, A., Lapedriza, A., Oliva, A., and Torralba, A.
\newblock Learning deep features for discriminative localization.
\newblock In \emph{Proceedings of the IEEE Conference on Computer Vision and
  Pattern Recognition}, pp.\  2921--2929, 2016.

\end{thebibliography}
\bibliographystyle{icml2019}

\end{document}